\documentclass[10pt]{article} % For LaTeX2e
\usepackage[preprint]{tmlr}

\usepackage{amsmath,amsfonts,bm}

\def\eqref#1{equation~\ref{#1}}
\def\1{\bm{1}}

\DeclareMathAlphabet{\mathsfit}{\encodingdefault}{\sfdefault}{m}{sl}
\SetMathAlphabet{\mathsfit}{bold}{\encodingdefault}{\sfdefault}{bx}{n}

\usepackage{hyperref}
\usepackage{url}
\usepackage[utf8]{inputenc}          % UTF-8 source → pdflatex
\usepackage[T1]{fontenc}             % 8-bit output font encoding
\usepackage{lmodern}                 % Latin Modern Type-1 fonts
\usepackage[whole]{bxcjkjatype}      % one-line CJK enablement
\usepackage{placeins}                % adds \FloatBarrier
\usepackage{flafter}                 % prevents floats before reference
\usepackage{authblk}
\usepackage{amsmath}                 % \operatorname, \dfrac, \text, …
\usepackage{amssymb}                 % \mathbb, \mathcal, \mathfrak, …
\usepackage{amsfonts}
\usepackage{booktabs}                % \toprule, \midrule, …
\usepackage{multirow}
\usepackage{graphicx}
\usepackage{setspace}
\usepackage{microtype}

\include{dataset}
\include{brier}
\include{brier_difference}
\include{ece}
\include{profit}
\include{confidence}
\newcommand{\var}[1]{\csname #1\endcsname}

\title{Forecasting Supply Chain Disruptions with Foresight Learning}

\author{\name Benjamin Turtel\thanks{Contact: ben@lightningrod.ai},  Paul Wilczewski, Kris Skotheim \\ \centering \addr \href{https://lightningrod.ai}{Lightning Rod Labs} }

\begin{document}

\maketitle

\begin{abstract}

Anticipating supply chain disruptions before they materialize is a core challenge for firms and policymakers alike. A key difficulty is learning to reason reliably about infrequent, high-impact events from noisy and unstructured inputs - a setting where general-purpose models struggle without task-specific adaptation. We introduce an end-to-end framework that trains LLMs to produce calibrated probabilistic forecasts using realized disruption outcomes as supervision. The resulting model substantially outperforms strong baselines - including GPT-5 - on accuracy, calibration, and precision. We also show that training induces more structured and reliable probabilistic reasoning without explicit prompting. These results suggest a general pathway for training domain-specific forecasting models that produce decision-ready signals. To support transparency we open-source the evaluation dataset used in this study. \footnote{Dataset available at: \href{https://huggingface.co/datasets/LightningRodLabs/supply-chain-predictions}{https://huggingface.co/datasets/LightningRodLabs/supply-chain-predictions}}
\end{abstract}

\section{Introduction}

\label{sec:intro}Supply chain disruptions are costly and difficult to anticipate because conventional indicators are delayed, revised, or only partially informative. This creates a persistent forecasting gap: decisions must be made in real time, while reliable signals often arrive only after conditions have deteriorated. News, by contrast, provides a timely stream of information on geopolitical tensions, trade restrictions, labor disputes, and other developments that may precede measurable disruptions. Leveraging these signals requires incorporating large volumes of unstructured text into probabilistic forecasting.

We therefore study supply chain forecasting as a probabilistic prediction problem that uses recent news to estimate future disruptions. Given a window of recent news, we estimate the likelihood that a disruption index will experience a large increase in the following month. The core challenge is not only accessing this information, but extracting and aggregating predictive signals from unstructured, evolving text in a way that supports reliable probabilistic inference.

Unlike prior approaches that use LLMs primarily to extract structured signals from text before applying downstream models, we train the model end-to-end to directly produce calibrated probabilistic forecasts. This formulation allows the model to jointly (i) identify salient signals in unstructured inputs, (ii) reason over those signals in a domain-adaptive manner, and (iii) produce likelihood estimates that are aligned with observed outcomes.

Our approach builds on Foresight Learning \cite{turtel2026}, a reinforcement learning framework for training models to produce probabilistic forecasts using supervision from realized outcomes. We extend this framework to time series measures of supply chain disruptions by constructing forecasting examples from timestamped news and realized disruption outcomes in a strictly forward-looking setting. We define a one-month-ahead task based on large month-over-month increases in disruption indexes.

We make four main contributions. First, we introduce a new forecasting task linking real-time news to future supply chain disruption events. Second, we develop an end-to-end modeling approach that trains LLMs directly on raw news inputs to produce probabilistic forecasts. Third, we show that this approach improves predictive performance, achieving lower Brier scores, reduced calibration error, and higher precision compared to pretrained models and baselines. Fourth, we demonstrate that training under a forecasting objective induces more structured and decision-relevant reasoning behavior, such as improved uncertainty handling and signal prioritization, without additional prompting.

\section{Related Work}
 
\subsection{Supply chain risk modeling}
 
Prior work uses unstructured text, e.g. news and social media, to monitor supply chain risk. Early approaches apply sentiment analysis, topic modeling, and supervised classification to identify disruptions and track risk signals. Recent surveys~\cite{gelastopoulos2025} emphasize that this literature focuses on detection and visibility rather than out-of-sample prediction.
 
A complementary line of work explores predictive modeling using structured data, network representations, or features derived from text. While these approaches can improve forecasting performance, they typically rely on structured inputs or task-specific features, limiting their ability to leverage raw textual signals.
 
Recent work also introduces aggregate disruption indices constructed from granular trade data~\cite{liu2023}, enabling systematic tracking of global supply chain stress. However, these indices are based on realized outcomes and do not incorporate information from text.
 
\subsection{Large language models for supply chains}
 
Recent work explores large language models (LLMs) in supply chain settings, primarily for information extraction, reasoning, and decision support. For example, hybrid frameworks such as SHIELD~\cite{cheng2024} use LLMs to extract structured signals from text, which are then fed into downstream predictive models. These approaches show that LLMs are effective at organizing complex textual information, but predictive performance is strongest when combined with statistical or graph-based models. Broader surveys~\cite{wang2025} similarly emphasize reasoning and planning rather than forecasting.
 
Related work also uses LLMs to construct text-based measures of supply chain risk from firm disclosures or managerial communication~\cite{fan2025}, but these approaches are largely descriptive or cross-sectional. Overall, relatively little work studies end-to-end prediction of time-varying disruptions using LLMs conditioned directly on raw text.
 
\subsection{Foresight Learning}
 
Our approach builds on Foresight Learning~\cite{turtel2026}, which formulates forecasting as supervised learning over temporally aligned text and future outcomes. Models are trained on examples constructed using only information available at the prediction time, mapping past information to probabilistic predictions of future events.
 
We adopt this framework in a domain-specific setting involving multi-entity news inputs and noisy disruption outcomes, focusing on supply chain forecasting as a concrete and practically relevant application.

\section{Data and Problem Setup}
\subsection{Data Sources}
\label{sources}

Our analysis combines two primary data sources: (i) timestamped news articles describing developments relevant to global supply chains, and (ii) quantitative measures of supply chain disruptions constructed from trade data.
For disruption measurement, we use supply chain disruption indexes\footnote{\href{www.disruptions.supply}{Supply Disruptions Index}} introduced in recent work \cite{liu2023}, which leverage granular international trade data to produce high-frequency indicators of supply chain stress. These indexes provide monthly measures of disruption intensity at both the country and product levels.

We pair these indexes with a corpus of news articles covering logistics, manufacturing, trade policy, commodities, and geopolitical developments. Each article is timestamped and can be associated with specific countries and products.

Together, these data sources enable the construction of forecasting problems that link recent news to future disruption outcomes.

\subsection{Forecasting Questions}

Each data point in our dataset is a forecasting question: a natural-language, entity-specific, time-indexed prediction about future supply chain disruptions.
A forecasting question is defined by:

\begin{itemize}
    \item an entity e (country or product),
    \item a prediction month t (the month at which the prediction is made),
    \item a news context consisting of articles available up to month t,
    \item current value for the index at t for entity e,
    \item a binary outcome indicating whether a disruption event occurs in month t+1.
\end{itemize}

Each question is formulated as a natural-language prompt that specifies the current state of the disruption index, a precise event definition, and resolution criteria. The model is asked to estimate the probability that a disruption event will occur in the following month.

A representative example is shown below:

As of October 2025, the supply chain disruption index for furniture is 0.53, having increased by 0.20 from the previous month. Will there be a supply chain shock for furniture next month? A shock is defined as a month-over-month increase exceeding 1 standard deviation (0.35) of historical changes.

The full prompt template, including news context and answer format, is provided in Appendix \ref{app:prompt-template}.

\subsection{News Context Construction}

For each forecasting question (e,t), we construct a news context by retrieving a set of recent articles relevant to the entity and its supply chain environment, restricted to those published at or before month t.

News articles are obtained via time-aware search over publicly available sources. Given the entity and prediction month, we identify and include articles describing developments in logistics, manufacturing, trade policy, commodities, and geopolitics that may affect the entity’s supply chain conditions.

This approach provides a concise and up-to-date context reflecting information available at the prediction month, ensuring the task is free of look-ahead bias.

\subsection{Event Definition and Labeling}

We define disruption events based on changes in the supply chain disruption index introduced in Section \ref{sources}.

Let $I_{e,t}$ denote the disruption index for entity e at month t, where e may correspond to either a country or a product. We define the binary outcome:

\begin{equation}
    y_{e,t+1} = \mathbf{1}\left(I_{e,t+1} - I_{e,t} \geq \sigma_e\right)
\end{equation}

where $\sigma_e$​ is the standard deviation of month-over-month changes in the disruption index for entity e, estimated from training data.

This definition identifies disruption events as unusually large increases in disruption intensity, corresponding to a meaningful deterioration in supply chain conditions. Scaling the threshold by $\sigma_e$ accounts for differences in volatility across entities, allowing for a consistent definition of “extreme” disruptions.

All labels are computed from future realizations of the disruption index relative to the prediction time t, ensuring the setup avoids look-ahead bias. Because the threshold $\sigma_e$ is estimated using only training data, the labeling procedure does not incorporate information from future periods.

\subsection{Prediction Task}

The resulting task is to estimate:

\begin{equation}
    P(y_{e,t+1} = 1 \mid \text{news}_{\le t}, e)
\end{equation}

the probability that entity e experiences a disruption event in month t+1, given only information available at time t.

We frame this as a probabilistic forecasting problem in which the model produces both a numerical probability and a natural-language explanation grounded in the news context. During training, these probability estimates are evaluated against realized outcomes and used as rewards for optimization.

\subsection{Dataset Statistics and Splits}
\label{dataset_stats}

We construct forecasting questions for entity–month pairs with both news context and future disruption outcomes available, restricting the sample to observations from 2022 onward. Although earlier data is available, we focus on the post-2022 period to avoid the extreme volatility associated with the COVID-19 shock and to study supply chain dynamics in a more stable regime.

Table 1 summarizes the resulting dataset, including the number of countries and products, the time span covered, the total number of forecasting questions, and the overall disruption event rate.

\begin{table}[htbp]
\centering
\begin{tabular}{lcc}
\hline
Metric & Training & Test \\
\hline
Time span & 1/2022 - 9/2025 & 10/2025 - 1/2026 \\
\# Countries & 25 & 25 \\
\# Products & 88 & 88 \\
\# Forecasting questions & 4,972 & 452 \\
Event rate (\%) & 14.9\% & 10.4\% \\
\hline
\end{tabular}
\caption{Dataset Summary Statistics}
\end{table}

We partition the dataset chronologically by prediction month, using January 2022 through September 2025 for training and October 2025 through January 2026 for testing. This ensures that all training examples precede the test period, preventing information leakage and reflecting a realistic forecasting setting.

\section{Model and Training}

\subsection{Learning Framework}

We frame supply chain disruption prediction as a probabilistic forecasting task in which supervision is provided by future realizations of disruption events. For each forecasting question defined by entity e and time t, the model observes only information available up to t and predicts whether a disruption event will occur at t+1.

We adopt the Foresight Learning framework \cite{turtel2026}, which enforces temporal information constraints by constructing inputs from a masked information state excluding post-t data, while labels are derived from future outcomes. This preserves the causal structure of forecasting in offline training. Within this setup, the model is trained to produce probabilistic forecasts using realized disruption events.

\subsection{Base Model}

Our base model is GPT-OSS-120B, an open-source 120B-parameter decoder-only transformer pretrained on general-domain and technical text, chosen as a strong general-purpose model.

We adapt the model using Low-Rank Adaptation (LoRA) with rank 32, enabling efficient fine-tuning with a small number of trainable parameters while keeping the base model fixed. This allows specialization to the forecasting task without full-parameter updates.

\subsection{Model Inputs}

Each input consists of several components:

\begin{itemize}
    \item Current and prior index values: the current and prior months' index values for the entity.
    \item News context: recent articles associated with entity e, restricted to information available at or before time t, represented as raw text, summaries, or headlines.
    \item Forecasting prompt: a natural-language query specifying the entity, current disruption conditions, and the event definition.
\end{itemize}

These components are concatenated using a prompt template that instructs the model to produce a probabilistic forecast. All inputs fit within the model's context window, with no truncation or sliding-window mechanisms required.

\subsection{Training Objective and Optimization}

The model outputs a probability $p_{e,t+1} \in [0,1]$, representing the predicted likelihood that entity $e$ will experience a disruption event at time $t+1$. We optimize the model using a GRPO-style reinforcement learning objective, where rewards are defined by the log score under the realized outcome. For prediction $p_{e,t+1}$ and outcome $y_{e,t+1}$, the reward is:

\begin{equation}
    r = y_{e,t+1} \log p_{e,t+1} + (1 - y_{e,t+1}) \log(1 - p_{e,t+1})
\end{equation}

This objective incentivizes accurate and well-calibrated probabilistic forecasts, as maximizing expected reward corresponds to maximizing likelihood under the true data-generating process. It also enables optimization over full model outputs, including both probabilities and associated 
reasoning.

Optimization is performed over LoRA parameters using standard policy optimization methods, with base model weights frozen. Reported results correspond to the final converged checkpoint.

\section{Results}

\subsection{Baselines}
We compare against the following baselines:
\begin{itemize}
\item \textbf{Historical baseline rate:} predicts the training-sample average disruption rate.
\item \textbf{Untrained language model:} a general-purpose LLM used in a prompted setting to produce forecasts from news context without task-specific training.
\item \textbf{Trained language model:} the same model fine-tuned using the proposed framework.
\end{itemize}
These comparisons isolate the value of conditioning on news and of task-specific training.

\subsection{Evaluation}
We evaluate model performance on held-out forecasting questions. The held-out 
sample is strictly temporally subsequent to the training data (see 
Section~\ref{dataset_stats}), ensuring no information leakage. All models are 
evaluated using the same input context and prompt template, and we include a 
historical baseline that predicts a constant probability equal to the empirical 
disruption rate in the training data.

Performance is measured using standard probabilistic metrics: Brier score, 
Brier skill score (BSS), expected calibration error (ECE), and 
Precision@10\%. The Brier score measures the accuracy of probabilistic 
predictions, while BSS, as measured relative to the historical baseline, 
captures the percentage improvement in Brier score, with positive values 
indicating better performance. Precision@10\% is defined as the precision 
among the top 10\% of predictions ranked by predicted probability, i.e., the 
fraction of highest-confidence predictions that correspond to true disruption 
events.

ECE measures calibration, i.e., how closely predicted probabilities align with 
realized event frequencies. Lower Brier score and ECE therefore indicate better 
predictive accuracy and more reliable probability estimates, while higher BSS 
and Precision@10\% indicate better performance. Precision@10\% is particularly 
relevant in practice, as decision-makers typically act on a limited set of 
high-confidence alerts rather than the full predicted distribution.

\subsection{Aggregate Probabilistic Performance}

Table 2 summarizes performance on the test set. The fine-tuned model outperforms the pretrained base model, the historical baseline, and a frontier general-purpose model across all reported metrics, with substantial gains in both accuracy and calibration. In particular, the fine-tuned model produces probabilities that more closely match observed disruption frequencies, improving the reliability of its forecasts; relative to the pretrained model, ECE decreases by nearly 70\%.

In addition, the fine-tuned model achieves markedly higher precision@10\%, indicating that its highest-confidence predictions are significantly more likely to correspond to true disruption events. This suggests improved ranking quality and practical utility in top-k decision settings compared to all baselines.

\begin{table}[htbp]
\centering
\begin{tabular}{lcccc}
\hline
Model & Brier~$\downarrow$ & Brier Skill Score~$\uparrow$ & ECE~$\downarrow$ & Precision~$\uparrow$ \\
\hline
Trained model & 0.0791 & 16.9\% & 0.0525 & 0.3478 \\
GPT-5 & 0.1203 & $-26.4$\% & 0.1304 & 0.0870 \\
GPT-OSS-120B & 0.1433 & $-50.5$\% & 0.1740 & 0.1304 \\
Historical Baseline & 0.0952 & 0\% & --- & --- \\
\hline
\end{tabular}
\caption{Model performance on the held-out test set. Lower Brier score and ECE indicate better performance, while higher Brier skill score and precision are better.}
\label{tab:model-performance}
\end{table}

The fine-tuned model outperforms both the pretrained base model, the naive baseline, and a frontier general-purpose model across all metrics. Fine-tuning yields improvements in both probabilistic accuracy and calibration. Relative to the pretrained base model, the reduction in ECE represents a nearly 80\% reduction in calibration error.

\begin{figure}[htbp]         
  \centering
  \includegraphics[width=\linewidth]{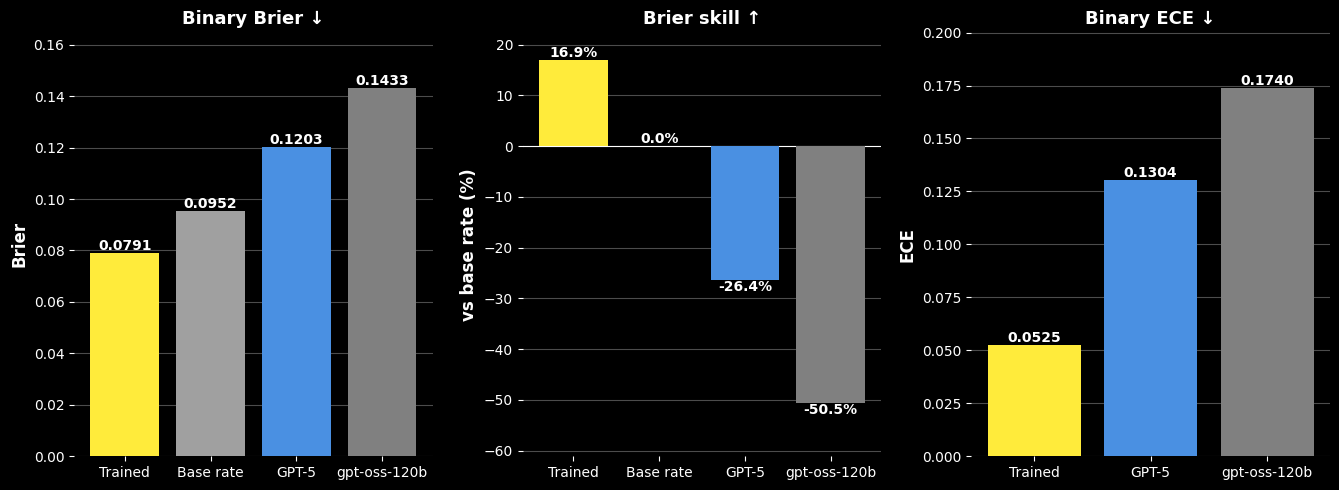}
  \caption{Aggregate performance on the held-out test set}
  \label{fig:aggregate-performance}
\end{figure}

\subsection{Calibration}

Calibration improves substantially after training. ECE decreases from 0.1740 for the pretrained model to 0.0525, indicating closer alignment between predicted probabilities and realized outcomes. The historical baseline, which predicts a constant rate, has no discriminative power.

Figure 2 shows a reliability diagram: predicted probabilities from the trained model closely track empirical frequencies, with higher predicted risk corresponding to higher observed disruption rates.

\begin{figure}[htbp]         
  \centering
  \includegraphics[width=0.5\linewidth]{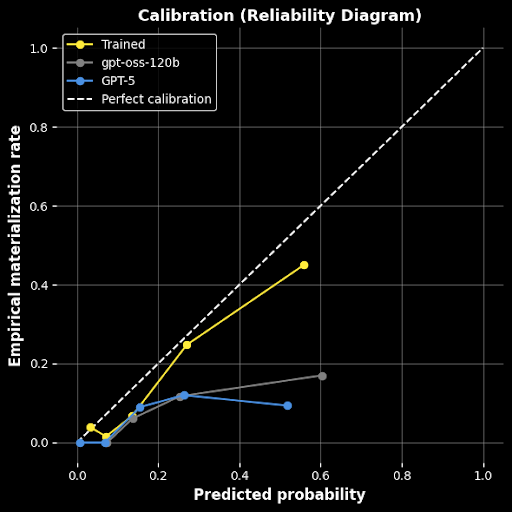}
  \caption{Reliability diagram on the test set showing empirical disruption rates as a function of predicted disruption probabilities.}
  \label{fig:calibration}
\end{figure}

\section{Discussion}
\subsection{Interpretation of Results}

Our results show that large language models trained under a foresight-oriented objective can extract predictive signals from news and produce reliable probabilistic forecasts of supply chain disruptions. This has several implications.

First, gains over historical baselines and prompted LLMs highlight the importance of task-specific adaptation. While general-purpose LLMs excel at language understanding, they do not naturally produce well-calibrated forecasts in out-of-domain settings. Fine-tuning under a forecasting objective improves both discrimination and alignment with realized outcomes, indicating that predictive reasoning can be strengthened through targeted supervision.

Second, the results emphasize the value of news as an anticipatory signal source. Traditional indicators are inherently lagging, whereas news captures early developments—such as policy shifts, labor disputes, and geopolitical tensions—that may precede disruptions. The model's ability to leverage this information without feature engineering suggests that LLMs can flexibly aggregate heterogeneous textual signals.

Third, improvements in calibration are particularly important for decision-making. Many applications require probabilities that reflect true event frequencies, not just relative risk rankings. The reduction in calibration error indicates that the trained model produces probability estimates that are more reliable for downstream use.

These results suggest that textual data can support supply chain risk assessment when paired with appropriate training frameworks. More broadly, they highlight the potential for LLMs to generate reliable probabilistic predictions from complex information.

\subsection{Improvements in Reasoning Behavior}
\label{reasoning-improvements}

Beyond calibration and aggregate performance, Foresight Learning leads to systematic improvements in domain reasoning. We observe consistent differences in reasoning between pretrained and fine-tuned models on identical forecasting questions. The pretrained model tends to produce descriptive summaries of recent events, often emphasizing salient but backward-looking information without clearly linking them to future outcomes. In contrast, the fine-tuned model exhibits more structured and forward-looking reasoning, more consistently identifying developments that plausibly propagate into future disruptions and explicitly connecting them to the event definition.

\begin{table}[htbp]
\centering
\begin{tabular}{lp{5cm}p{5cm}}
\hline
Dimension & Pretrained Model & Fine-tuned Model \\
\hline
Structure & Short, largely heuristic reasoning & Multi-step reasoning with explicit intermediate structure \\[6pt]
Temporal orientation & Describes recent events & Connects current signals to future outcomes \\[6pt]
Quantitative grounding & Rare use of numerical reasoning & Frequent use of thresholds, distributions, and rough calculations \\[6pt]
Base rate usage & Implicit or absent & Explicit anchoring to baseline probabilities \\[6pt]
Uncertainty handling & Single-pass conclusions & Iterative refinement with intermediate estimates \\[6pt]
Use of news & Summarizes salient events & Selectively incorporates signals relevant to disruption risk \\
\hline
\end{tabular}
\caption{Systematic differences in reasoning behavior before and after training.}
\label{tab:reasoning-behavior}
\end{table}

These qualitative differences are also reflected in the frequency of specific reasoning behaviors. Using an automated evaluator (Appendix \ref{app:automated-eval}), we measure the presence of several forms of probabilistic reasoning across the evaluation set. As shown in Table 4, the fine-tuned model exhibits large increases in behaviors associated with structured probabilistic reasoning, including base-rate anchoring, explicit forecasting models, and iterative refinement of uncertainty.

\begin{table}[htbp]
\centering
\begin{tabular}{lcc}
\hline
Behavior & Pretrained & Fine-tuned \\
\hline
Base rate anchoring & 0.09 & 0.50 \\
Statistical modeling & 0.48 & 1.00 \\
Explicit forecasting model & 0.25 & 0.96 \\
Evidence $\rightarrow$ outcome linkage & 0.67 & 0.70 \\
Probabilistic synthesis & 0.94 & 1.00 \\
Uncertainty refinement & 0.33 & 1.00 \\
\hline
\end{tabular}
\caption{Frequency of reasoning behaviors detected by automated evaluator. Values represent the fraction of responses exhibiting each behavior.}
\label{tab:reasoning-frequency}
\end{table}

Aggregating across behaviors, the average rubric score increases from 2.76 to 5.17 (out of 6), indicating a substantial shift toward more structured and quantitatively grounded reasoning.

These results highlight several systematic changes after training. First, the fine-tuned model more consistently adopts an explicit probabilistic framing, often invoking simple statistical assumptions (e.g., volatility, tail events) and informal forecasting models such as persistence or mean reversion. Second, it more clearly separates baseline probabilities from conditional adjustments, anchoring predictions to an unconditional likelihood before incorporating recent developments. Third, the model more frequently performs iterative refinement, revising intermediate estimates as additional factors are considered. Finally, it makes more consistent use of evidence by linking specific developments in the news to their expected impact on future disruption outcomes, rather than merely summarizing recent events.

Notably, these behaviors are not explicitly programmed or prompted, but emerge through training under the forecasting objective. The model learns to construct probabilistic reasoning strategies - such as baseline anchoring, simple stochastic modeling, and iterative updating - through exposure to temporally aligned prediction tasks.

These findings suggest that foresight-oriented training improves not only predictive accuracy but also the structure of probabilistic reasoning, enabling the model to more consistently translate textual information into calibrated forecasts. A representative example is shown in Appendix \ref{app:reasoning-comparison}.

\subsection{Limitations}

A key limitation is the noisy and incomplete relationship between news and realized disruptions, which places an upper bound on predictive performance. Some relevant developments may not be reported, while others may not translate into measurable disruptions.

Our formulation also focuses on one-month-ahead binary events, which abstracts from richer dynamics such as persistence and multi-period risk accumulation. Extending the framework to longer horizons or continuous outcomes would provide a more complete characterization of supply chain risk.

Finally, both training and evaluation are limited to the post-2022 period, when disruption levels remain elevated relative to pre-pandemic norms. This confines the analysis to a single regime, whereas supply chain dynamics may shift over time. Evaluating robustness to such changes in underlying distributions is an important direction for future work.

\section{Conclusion}

This paper studies the problem of forecasting supply chain disruptions from news using large language models trained under a foresight learning framework. We construct a temporally aligned dataset linking recent news to future disruption events and train an LLM to produce well-calibrated probabilistic forecasts.

Our results show that fine-tuned LLMs substantially outperform strong baselines in both predictive accuracy and calibration, demonstrating that signals in textual data can be effectively leveraged for supply chain forecasting. In particular, improvements in calibration suggest that predicted probabilities more closely reflect realized outcomes, an important property for forecasting under uncertainty. These findings contribute to a growing body of work suggesting that LLMs can move beyond descriptive and reasoning tasks toward quantitative prediction under uncertainty.

More broadly, this work highlights the potential of combining textual data with temporally consistent training objectives to address real-world forecasting problems. Future research directions include extending the framework to multi-horizon and multi-event forecasting, integrating additional data sources such as structured trade flows or firm-level disclosures, and improving interpretability and robustness under distributional shifts.

These findings show that LLMs trained under a foresight learning framework can produce calibrated, forward-looking forecasts of supply chain disruption risk from unstructured news.

\appendix

\section{Forecasting Prompt Template}
\label{app:prompt-template}

We construct each forecasting example as a structured prompt combining a prediction task, relevant context, and a standardized output format. A representative example is shown below.

\subsection*{Prompt Structure (Example)}

\begin{quote}
\textbf{Instruction:}
You are a supply chain analyst forecasting disruption shocks for specific trade flows (countries or products). Given the current disruption index and relevant news, estimate the probability of a disruption next month.

\textbf{Question:}
As of October 2025, the supply chain disruption index for \texttt{residues\_waste} is 0.69, having decreased by 0.26 from the previous month. Will there be a supply chain shock for \texttt{residues\_waste} next month? A shock is defined as a month-over-month increase exceeding 1 standard deviation (0.46) of historical changes.

\textbf{Resolution Criteria:}
Resolves \textsc{Yes} if the disruption index increases by more than 0.46 from October 2025 to November 2025. Resolves \textsc{No} otherwise.

\textbf{Context (abridged):}
Recent news articles relevant to this question include: SAF supply shortfall projections indicating long-term deficits in aviation fuel supply due to limited biomass and scaling challenges; growth in agricultural waste management driven by increasing waste volumes and infrastructure constraints; expansion of waste-to-energy initiatives in Southeast Asia; and environmental disruptions such as flooding affecting waste management infrastructure. Full article texts are omitted for brevity.

\textbf{Output Format:}
Provide a probability between 0 and 1 representing the likelihood of a disruption event and a brief explanation grounded in the news context. Return the probability in the format \texttt{<answer>p</answer>} where $p \in [0, 1]$.

\textbf{Notes:}
All inputs are constructed using only information available at the prediction time. News context is summarized and filtered for relevance. Entities include both countries and product categories. Event thresholds are computed from historical training data.
\end{quote}

% -------------------------------------------------------

\section{Model Reasoning Before and After Training}
\label{app:reasoning-comparison}

We compare model outputs on an identical forecasting question (following the prompt structure described in Appendix~\ref{app:prompt-template}) before and after training. Formatting has been simplified for readability; reasoning is otherwise unchanged. Ellipses (\texttt{[...]}) indicate omitted segments. Bracketed annotations highlight key reasoning patterns described in Section~\ref{reasoning-improvements}.

\subsection*{Example Reasoning Before Training}

\begin{quote}
We need to decide if next month (December 2025) will see an increase $>0.46$ from current index 0.97. Currently increased $+0.28$ from previous month (October). To exceed 0.46 increase, need index to go to at least $0.97 + 0.46 = 1.43$. That's a large jump.

We need to consider news: Many articles about increased feedstock demand, innovations, but also some concerns about supply crunch. But does that imply disruption index will increase more? The disruption index likely measures supply chain disruptions. Current increase is $+0.28$, which is already above baseline. The news suggests both pressure (feedstock crunch) could increase disruptions, but also new technologies may alleviate. However a $>0.46$ jump seems large. Likely not.

Probability of Yes maybe low, like 0.2. Provide answer.

\medskip
\textbf{Final Answer:} \texttt{<answer>0.2</answer>}
\end{quote}

\subsection*{Example Reasoning After Training}

\begin{quote}
We need to estimate probability that next month will see a supply chain disruption shock for \texttt{residues\_waste}. The definition: a month-over-month increase exceeding 1 standard deviation of historical changes: $\mathrm{SD} = 0.46$. So if disruption index increases by $>0.46$ from 0.97 to $>1.43$, that would be a shock. But we need probability that increase $> 0.46$.

\medskip
\noindent\textit{[Structured setup]}~~Given current change from previous month was $+0.28$ (increase). That's less than 0.46. \texttt{[\ldots]} The question: ``Will there be a supply chain shock for \texttt{residues\_waste} next month?'' Shock defined as month-over-month increase exceeding 0.46. \texttt{[\ldots]}

\medskip
\noindent\textit{[Quantitative reasoning]}~~Alternative: use random walk: expected change = last change (0.28). So forecast next change $= 0.28$ with same volatility 0.46. Probability $> 0.46 ={}$? Under normal with mean 0.28, sd 0.46, probability that change $> 0.46$ is $= 0.18/0.46 \approx 0.3913$. So $Z > 0.3913$ gives probability $\approx 0.35$. \texttt{[\ldots]}

\medskip
\noindent\textit{[Base rate comparison]}~~Alternatively one may assume changes are zero-mean, standard deviation 0.46. Then probability change $> 0.46$ is about 0.16. \texttt{[\ldots]}

\medskip
\noindent\textit{[Self-correction / updating]}~~Thus probability maybe around 0.3. \texttt{[\ldots]} Thus likely probability maybe $0.25$--$0.3$. \texttt{[\ldots]}

\medskip
\noindent\textit{[Incorporating news signals]}~~Now incorporate news. Articles relate to waste and residues used for biofuel feedstock \texttt{[\ldots]} likely increase demand and pressure, possibly higher disruption\texttt{[\ldots]} Mixed \texttt{[\ldots]}

\medskip
\noindent\textit{[Final synthesis]}~~Thus, while upward pressure exists, a change exceeding 0.46 remains a tail event. Balancing trend, volatility, and mixed signals, I'll assign 0.30.

\medskip
\textbf{Final Answer:} \texttt{<answer>0.30</answer>}
\end{quote}

% -------------------------------------------------------

\section{Automated Evaluation of Reasoning Behavior}
\label{app:automated-eval}

To complement qualitative comparisons of model outputs, we use an automated evaluator to measure the presence of specific reasoning behaviors in model-generated forecasts. The goal of this analysis is not to assess overall reasoning quality, but to detect the use of concrete probabilistic reasoning patterns corresponding to the dimensions described in Section~6.2.

\subsection*{Evaluator Design}

We define a fixed rubric consisting of six binary reasoning behaviors:

\begin{enumerate}
    \item \textbf{Base-rate anchoring:} The reasoning explicitly references a baseline probability, typical frequency, or distributional prior (e.g., ``rare event,'' ``tail probability,'' or unconditional likelihood).

    \item \textbf{Statistical modeling:} The reasoning frames the problem in terms of a stochastic or distributional process (e.g., volatility, variance, or tail events).

    \item \textbf{Explicit forecasting model:} The reasoning specifies or assumes a model of temporal dynamics, such as persistence, trend continuation, or mean reversion.

    \item \textbf{Evidence--outcome linkage:} The reasoning connects specific pieces of evidence (e.g., news events) to their expected impact on future disruption outcomes.

    \item \textbf{Probabilistic synthesis:} The reasoning combines multiple factors (e.g., base rates, trends, and news) into a final probability estimate.

    \item \textbf{Uncertainty refinement:} The reasoning revises or updates an intermediate estimate based on additional considerations.
\end{enumerate}

Each behavior is annotated as present (1) or absent (0) for a given reasoning trace.

\subsection*{Evaluation Procedure}

We apply the evaluator to model outputs on the held-out test set. For each forecasting question, we collect reasoning traces from both the pretrained and fine-tuned models using identical inputs and prompt templates. Each reasoning trace is then independently evaluated using a large language model, which is instructed to detect the presence of each behavior according to the rubric above.

The evaluator operates with deterministic decoding (temperature 0) and returns structured JSON outputs containing binary indicators for each behavior. We aggregate results across all evaluation examples to compute the mean frequency of each behavior for each model.

\subsection*{Prompting and Output Format}

The evaluator is implemented via a structured prompt that presents the reasoning trace and requests binary annotations for each behavior. The model is instructed to be strict and to mark a behavior as present only if it is clearly and explicitly demonstrated.

A simplified version of the prompt is shown below:

\begin{quote}
\begin{verbatim}
You are evaluating reasoning traces from a forecasting model.
Your task is to detect whether specific probabilistic reasoning
behaviors are present.
Be strict and literal. Only mark a behavior as present if it is
clearly demonstrated.

Analyze the following reasoning trace:

Return JSON:
{
  "base_rate": 0 or 1,
  "statistical_model": 0 or 1,
  "explicit_forecasting_model": 0 or 1,
  "evidence_linkage": 0 or 1,
  "probabilistic_synthesis": 0 or 1,
  "uncertainty_refinement": 0 or 1
}
\end{verbatim}
\end{quote}

\subsection*{Interpretation and Limitations}

This evaluation provides a coarse but scalable measure of reasoning behavior. Because the evaluator relies on a language model, annotations may be imperfect and sensitive to phrasing. The rubric is designed to prioritize precision over recall, and the resulting metrics should be interpreted as indicative of relative differences between models rather than exact measurements of reasoning quality.

Importantly, the evaluator measures the presence of specific reasoning patterns rather than their correctness or effectiveness. As such, these metrics complement - but do not replace - quantitative evaluation of forecasting performance.

\bibliographystyle{tmlr}
\bibliography{Ref}

@article{gelastopoulos2025,
  title={A Systematic Review of Text Mining Analytics for Supply Chain Risk Management Using Online Data},
  author={Gelastopoulos, Georgios and Keramydas, Christos},
  journal={Supply Chain Analytics},
  year={2025},
  url={https://www.sciencedirect.com/science/article/pii/S2949863525000676}
}

@article{liu2023,
  title={Supply Disruptions Index ({SDI}): Data and Methodology},
  author={Liu, Ernest and Smirnyagin, Vladimir and Tsyvinski, Aleh},
  journal={SSRN Working Paper},
  year={2023},
  url={https://papers.ssrn.com/sol3/papers.cfm?abstract_id=4430231}
}

@inproceedings{cheng2024,
  title={{SHIELD}: {LLM}-Driven Schema Induction for Predictive Analytics in {EV} Battery Supply Chain Disruptions},
  author={Cheng, Zhi-Qi and Dong, Yifei and Shi, Aike and Liu, Wei and Hu, Yuzhi and O'Connor, Jason and Hauptmann, Alexander G. and Whitefoot, Kate},
  booktitle={Proceedings of the 2024 Conference on Empirical Methods in Natural Language Processing: Industry Track},
  pages={303--333},
  year={2024},
  address={Miami, Florida, US},
  publisher={Association for Computational Linguistics},
  doi={10.18653/v1/2024.emnlp-industry.24},
  url={https://aclanthology.org/2024.emnlp-industry.24/}
}

@article{wang2025,
  title={{LLMs} for Supply Chain Management},
  author={Wang, Haojie and Jiang, Jiuyun and Hong, L. Jeff and Jiang, Guangxin},
  journal={arXiv preprint arXiv:2505.18597},
  year={2025},
  url={https://arxiv.org/abs/2505.18597}
}

@article{fan2025,
  title={Measuring Firm-Level Supply Chain Risk Using a Generative Large Language Model},
  author={Fan, Siyu and Wu, Yifei and Yang, Ruochen},
  journal={Finance Research Letters},
  volume={77},
  year={2025},
  doi={10.1016/j.frl.2025.107111},
  url={https://www.sciencedirect.com/science/article/abs/pii/S1544612325003745}
}

@article{turtel2026,
  title={Future-as-Label: Scalable Supervision from Real-World Outcomes},
  author={Turtel, Benjamin and Wilczewski, Paul and Franklin, Danny and Skotheim, Kris},
  journal={arXiv preprint arXiv:2601.06336},
  year={2026},
  url={https://arxiv.org/abs/2601.06336}
}

\end{document}